\documentclass[journal]{IEEEtran}
\usepackage{amsmath,amsfonts}
\usepackage{algorithmic}
\usepackage{algorithm}
\usepackage{array}
\usepackage[caption=false,font=normalsize,labelfont=sf,textfont=sf]{subfig}
\usepackage{textcomp}
\usepackage{stfloats}
\usepackage{url}
\usepackage{verbatim}
\usepackage{graphicx}
\usepackage{cite}
\usepackage{color}
\usepackage{booktabs}
\usepackage{makecell}
\definecolor{NewGreen}{rgb}{.43,.67,.27}
\definecolor{NewOrange}{rgb}{.92,.49,.19}
\definecolor{NewBlue}{rgb}{.26,.44,.76}
\usepackage[colorlinks,bookmarksopen,bookmarksnumbered,citecolor=green, linkcolor=red, urlcolor=magenta]{hyperref}

\begin{document}

\title{Quality-guided Skin Tone Enhancement\\for Portrait Photography}

\author{Shiqi Gao\IEEEauthorrefmark{1}, Huiyu Duan\thanks{\IEEEauthorrefmark{1} Equal contribution.}\IEEEauthorrefmark{1}, Xinyue Li, Kang Fu, Yicong Peng, Qihang Xu, Yuanyuan Chang, \\Jia Wang, Xiongkuo Min\IEEEauthorrefmark{2},~\IEEEmembership{Member,~IEEE}, and Guangtao Zhai\IEEEauthorrefmark{2},~\IEEEmembership{Senior Member,~IEEE}\thanks{\IEEEauthorrefmark{2} Co-corresponding authors.}
\thanks{Shiqi Gao, Huiyu Duan, Xinyue Li, Kang Fu, Yicong Peng, Jia Wang, Xiongkuo Min and Guangtao Zhai are with the Institute of Image Communication and Network Engineering, Shanghai Jiao Tong University, Shanghai 200240, China (e-mail: shiqigao@sjtu.edu.cn; huiyuduan@sjtu.edu.cn; xinyueli@sjtu.edu.cn; fuk20-20@sjtu.edu.cn; jack-sparrow@sjtu.edu.cn; jiawang@sjtu.edu.cn; minxiongkuo@sjtu.edu.cn; zhaiguangtao@sjtu.edu.cn).}
\thanks{Qihang Xu and Yuanyuan Chang are with TRANSSION, shanghai 200240, China (e-mail: qihang.xu@transsion.com, yuanyuan.chang@transsion.com).}}


\maketitle

\begin{abstract}
In recent years, learning-based color and tone enhancement methods for photos have become increasingly popular.
However, most learning-based image enhancement methods just learn a mapping from one distribution to another based on one dataset, lacking the ability to adjust images continuously and controllably.
It is important to enable the learning-based enhancement models to adjust an image continuously, since in many cases we may want to get a slighter or stronger enhancement effect rather than one fixed adjusted result.
In this paper, we propose a quality-guided image enhancement paradigm that enables image enhancement models to learn the distribution of images with various quality ratings.
By learning this distribution, image enhancement models can associate image features with their corresponding perceptual qualities, which can be used to adjust images continuously according to different quality scores.
To validate the effectiveness of our proposed method, a subjective quality assessment experiment is first conducted, focusing on skin tone adjustment in portrait photography. 
Guided by the subjective quality ratings obtained from this experiment, our method can adjust the skin tone corresponding to different quality requirements. 
Furthermore, an experiment conducted on 10 natural raw images corroborates the effectiveness of our model in situations with fewer subjects and fewer shots, and also demonstrates its general applicability to natural images.
Our project page is \href{https://github.com/IntMeGroup/quality-guided-enhancement}{https://github.com/IntMeGroup/quality-guided-enhancement}.
\end{abstract}

\vspace{-1.0mm}

\vspace{-3.5mm}
\section{Introduction}
\vspace{-1.0mm}
\IEEEPARstart{D}{igital} 
photos taken in various uncontrolled environments may suffer from low dynamic ranges or distorted color tones \cite{kim2013optimized,lim2017contrast,duan2023masked,wang2023visual}.
Although several cascaded modules such as white balance, exposure compensation, hue or saturation adjustment, tone mapping and gamma correction, \textit{etc.}, are generally applied in digital cameras \cite{karaimer2016software,liang2021ppr10k}, which are manually tuned by experienced engineers, the output images may still need post-processing or retouching to further enhance the visual quality \cite{duan2022unified,duan2022develop,liang2021ppr10k}.
However, photo retouching is a complicated task and usually requires expertise in photography and professional processing softwares such as PhotoShop \cite{zeng2020learning}.
Moreover, users generally have diverse preferences for image aesthetics \cite{kim2020pienet}, thus for different users, the expected degree of image enhancement may be different.

With the advancement of machine learning techniques, many learning-based automatic photo enhancement methods have been proposed.
Bychkovsky \textit{et al.} \cite{bychkovsky2011learning} have constructed a MIT-Adobe FiveK dataset, which contains 5000 raw images with corresponding human-retouched ground-truth images.
Based on this dataset, some conventional machine learning-based methods have been proposed to enhance the quality of images \cite{bychkovsky2011learning,yan2014learning}.
Recently, benefiting from the advancement of deep learning, many deep neural network-based methods have been proposed for image enhancement \cite{gharbi2017deep,yan2016automatic,chen2018deep,park2018distort,ma2022low}.
Considering the computational and memory costs of these deep learning-based methods are heavy, some studies have also explored to use 3D lookup tables (LUTs) to conduct image enhancement \cite{zeng2020learning,yang2022adaint,yang2022seplut}, which is a more lightweight but effective method.
However, most of these methods just learn a mapping from one distribution to another distribution based on one dataset, while the output of an enhancement model for a given input is fixed.
Thus, it is hard to retouch the output based on these models to make it conform to the visual quality of various users.


\begin{figure}[!t]
  \centering
  \includegraphics[width=0.95\linewidth]{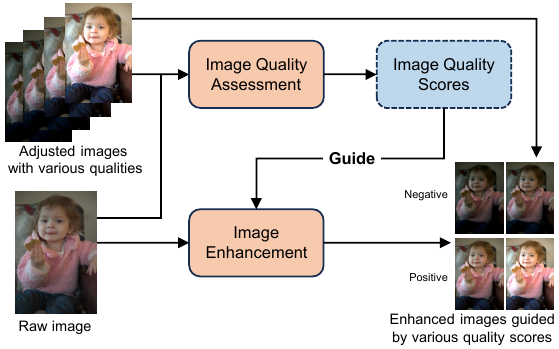}
  \vspace{-3.5mm}
  \caption{Quality-guided Image Enhancement Paradigm. Our framework first applies image quality assessment for a raw image and various adjusted images to obtain the quality scores for the adjusted images, then utilizes a raw image and the corresponding quality score of an adjusted image as the input for the enhancement network, and optimizes the network to make the output enhanced image as close as possible to the adjusted image. ``Positive'' and ``Negative'' represent better perceptual quality and worse perceptual quality compared to the raw image, respectively.}
  \vspace{-6.0mm}
  \label{Paradigm}
\end{figure}

Various image enhancement methods generate enhanced images with various qualities.
Many subjective and objective image quality assessment models have been proposed to evaluate the perceptual quality for low-light enhancement \cite{zhai2021perceptual} or haze image enhancement \cite{min2018objective,min2019quality}.
However, few studies have explored the quality assessment for image retouching \cite{gu2017learning,vu2012quality,min2024perceptual}.
One objective for image quality assessment is to introduce human opinion for enhancing images \cite{duan2022confusing}.
However, most current works only study the problem of assessing the performance of image enhancement models, while image enhancement methods based on various quality scores have rarely been studied. 

In this paper, we propose a novel quality-guided image enhancement framework that enables image enhancement models to adjust an image continuously and controllably.
Specifically, we first apply the proposed framework to 3D LUT image enhancement methods due to their feasibility in practical applications.
The proposed framework guides the 3D LUT image enhancement model to learn the distribution of images with various quality labels, and associate image features with their corresponding perceptual quality scores.
Therefore, during the inference process, the image enhancement model can adjust an image according to various quality scores.
To validate the effectiveness of the proposed framework, we conduct an experiment on the skin tone adjustment task for portrait photography since users usually would like to adjust the image skin tone slightly according to their own preferences.
We first construct a skin tone image quality assessment database (STIQAD) which includes 85 raw portrait images and 1105 adjusted face images with corresponding mean opinion scores (MOSs) collected from 20 subjects.
By predicting an adjusted image using the corresponding raw image and quality label, the image enhancement model can associate the quality score with the corresponding adjustment strategy and adjust skin tone according to different preferences.
Moreover, we further validate the effectiveness of our proposed method on natural images, and the experimental results also demonstrate that our framework can work well in situations with fewer subjects and fewer instances.
Extensive additional experiments demonstrate the generalization and adaptation ability of our framework.
The contributions are summarized as follows.

\begin{itemize}
    \item We propose a novel quality-guided image enhancement framework, which can control the image enhancement process according to quality scores.
    \item A skin tone image quality assessment database is established, which includes raw portrait images, their adjusted images and the corresponding quality scores of these adjusted images.
    \item We apply the proposed framework on the 3D LUT image enhancement method and conduct the quality-guided image enhancement on the STIQAD, and comprehensive experimental results demonstrate that our proposed framework can achieve continuous and controllable image enhancement. 
    \item Extensive additional experiments demonstrate the generalization and adaptation ability of our method, which include enhancing on natural images, replacing the subjective MOSs with objective IQA scores, using other image enhancement algorithm, performing user-specific enhancements, \textit{etc}.
\end{itemize}

\vspace{-1.5mm}
\section{Related Work}

\vspace{-1.5mm}
\subsection{Image Enhancement}
\vspace{-1.5mm}
Image enhancement aims at adjusting images to make them more conformed to human visual preferences.
Traditional methods mainly use manually extracted features to perform image enhancement.
Kim \textit{et al.} \cite{kim2012new} proposed to perform convolution through smoothing and sharpening operators to increase image clarity.
Van \textit{et al.} \cite{van2007edge} proposed to perform filter correction on the frequency domain for image enhancement.
Recently, many learning-based image enhancement methods have been proposed.
Gharbi \textit{et al.} \cite{gharbi2017deep} presented a CNN-based bilateral learning framework to predict the coefficients in bilateral space and then enhance the image quality.
Liu \textit{et al.} \cite{liu2020color} used a global parameter extractor subnetwork and a local feature extractor subnetwork to learn global and local features and then enhance images correspondingly.

\vspace{-3.5mm}
\subsection{Learnable Lookup Tables}
Recently, some works have studied using learnable 3D LUTs to perform image enhancement considering the high computational efficiency of the 3D LUT.
Zeng \textit{et al.} \cite{zeng2020learning} presented the first work of combining 3D LUTs and deep neural networks, which uses a lightweight CNN to predict the weights of multiple learnable 3D LUTs, and then performs image enhancement based on the fused LUT.
Wang \textit{et al.} \cite{wang2021real} replaced the lightweight CNN with a lightweight dual-head weight prediction network, which outputs a 1D weight vector for basic 3D LUT fusion and a 3D weight map for pixel-level category fusion.
Yang \textit{et al.} \cite{yang2022adaint} proposed to modify the uniform sampling in original 3D LUT to non-uniform sampling, and use lightweight CNN to predict non-uniform sampling coordinates.
Yang \textit{et al.} \cite{yang2022seplut} further proposed a separable adaptive lookup table method for real-time image enhancement, which separately predicts 1D and 3D LUTs for improving performance.

\vspace{-3.5mm}
\subsection{Quality Assessment for Enhanced Images}
Many works have studied the perceptual quality assessment \cite{duan2023attentive,duan2024quick,zhu2023perceptual,duan2018perceptual} for image enhancement.
Chen \textit{et al.} \cite{chen2014quality} conducted a subjective quality assessment experiment by comparing two enhanced results, and proposed a no-reference IQA model.
Gu \textit{et al.} \cite{gu2017learning} proposed a no-reference image quality assessment model for enhanced images with big data.
Min \textit{et al.} \cite{min2018objective} conducted a double-stimulus subjective quality assessment experiment for dehazed images and proposed an objective quality assessment model.
Zhai \textit{et al.} \cite{zhai2021perceptual} constructed a low-light enhancement image quality assessment database, and proposed a full-reference low-light enhancement quality assessment model.
However, most of these quality assessment works only consider performing IQA for enhanced images, while image enhancement based on quality assessment has rarely been studied.

\vspace{-2.5mm}
\section{Skin Tone Image Quality assessment Database (STIQAD)} \label{section:sec3}

We first construct a skin tone image quality assessment database to address the absence of a corresponding image retouching quality assessment database.
Skin tone plays a significant role for the perceived quality of portrait images captured by smart devices or cameras \cite{tu2022end,duan2022saliency,tu2022iwin}.
The purpose of this database is to associate the various skin tones and the corresponding mean perceptual qualities, which can be used to facilitate image enhancement models to learn the quality distribution of various skin colors. 
The detailed process of constructing the STIQAD is introduced as follows.

\vspace{-3.7mm}
\subsection{Image Selection and Processing}
We collect 85 portrait images covering a wide range of skin tones and apply various color adjustments to these images.
Specifically, we first convert the pixel colors from the RGB space to the CIELAB space, and randomly adjust the $A$ and $B$ values in a specific range while keeping the $L$ value unchanged, since we mainly focus on the tone adjustment rather than the lightness adjustment.
This step generates a large number of adjusted images with various tones, but most adjusted images are of low perceived quality.
Then we manually screen these adjusted images to balance the low-quality and high-quality adjusted samples for each raw stimuli \cite{zhou2021omnidirectional}.
During the process of screening, we remove samples with similar visual appearances to ensure diversity in the dataset.
Typically, the low-quality samples we retained exhibit varying degrees of different color casts.
It should be noted that some over-tuned samples are discarded though some subjects may give them high-quality scores, since various subjects may have large differences in scoring these images \cite{wang2023aigciqa2023,zhu2022blind,cao2021debiased}, which may affect the performance of quality guidance \cite{wang2024understanding,yang2024aigcoiqa2024}.
As a result, we acquire 13 adjusted images for each raw portrait image, and about half of these adjusted images have lower perceptual quality and another half have higher perceptual quality compared to the raw image.
Overall, we acquire 1105 adjusted face images corresponding to 85 raw portrait images.
Since the background part of the portrait image may interfere the judgment of skin tone quality \cite{zheng2022decoupled,duan2019visual}, we parse the face part separately, and only skin and facial areas are shown to the subjects in the subjective experiments as discussed follows.

\begin{figure}
  \centering
  \includegraphics[width=0.91\linewidth]{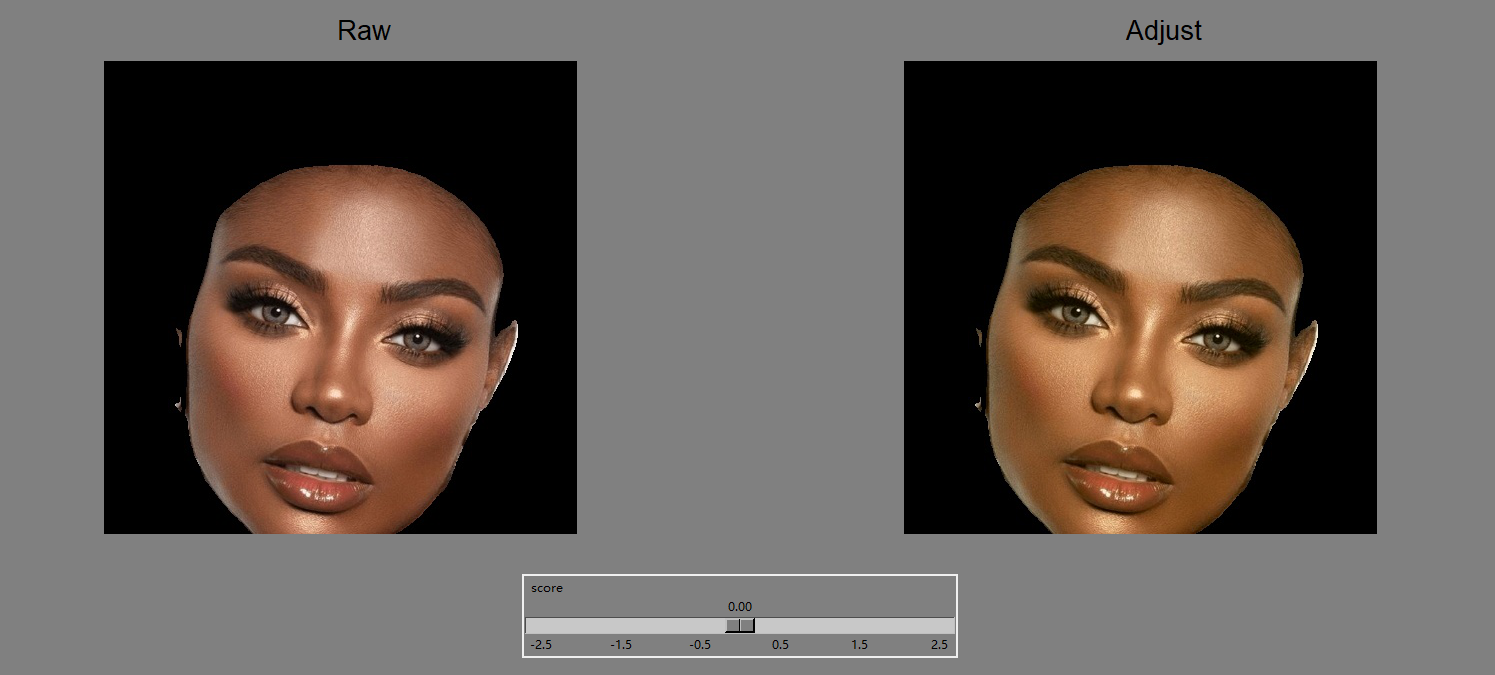}
  \vspace{-2.5mm}
  \caption{An illustration of the user interface of the subjective skin tone quality assessment experiment.}
  \vspace{-5.5mm}
  \label{interface}
\end{figure}

\vspace{-3.7mm}
\subsection{Psychophysical Experimental Setup}
{\bfseries\itshape Experimental environment and apparatus.} 
To mitigate the impact of different display parameters and ambient brightness on the  viewing experience of the subjects, the experiment is conducted in a room illuminated solely by the monitor. 
We use a 32-inch LCD monitor to conduct the experiment.
The color temperature of the display is set to 6,500K, and the brightness is adjusted to 50 \cite{duan2019dataset,zhou2019no}. 
Subjects are seated approximately 1.5 m away from the monitor, and maintain this distance throughout the experiment. 
This arrangement allows the subject to look at the screen without being too close, allowing the simultaneous comparison of the left and right images. 
To minimize color shifts caused by viewing the display from different angles, subjects are instructed to align their heads approximately at the center of the screen.

{\bfseries\itshape Experimental interactive interface.} Fig. \ref{interface} shows the user interface of our subjective skin tone quality experiment. 
The left side displays the raw portrait photograph, while the right side shows the adjusted image. 
Subjects are asked to evaluate the quality of the adjusted image, by giving the quality rating of how much the adjusted image is improved or decreased relative to the original image.
Unlike the 0-5 scale that is usually used in previous IQA studies \cite{duan2022confusing}, our subjective experiment allows the score bar to be dragged continuously between -2.5 and 2.5 points, where a negative value means the adjusted image has a lower quality compared to the raw image and a positive value represents the adjusted image has a higher quality.
This method is also a five-level scale with 1 score representing one level, and each level from -2.5 to 2.5 means much worse, worse, similar, better, and much better, respectively.

{\bfseries \itshape Experimental procedure.} As suggested by \cite{series2012methodology}, a minimum of 15 subjects is required in a subjective experiment to ensure reliable results. 
In this work, we recruit 20 experts who have expertise in photography and evaluation.
Each participant has normal or corrected-to-normal vision and color recognition. 
Before scoring, we display 20 pairs of facial images for training, including various degrees of good and bad adjustment cases. 
During the experiment, subjects are instructed to take breaks every 20 minutes to prevent excessive fatigue.

\begin{figure}[t]
  \centering
  \includegraphics[width=0.9\linewidth]{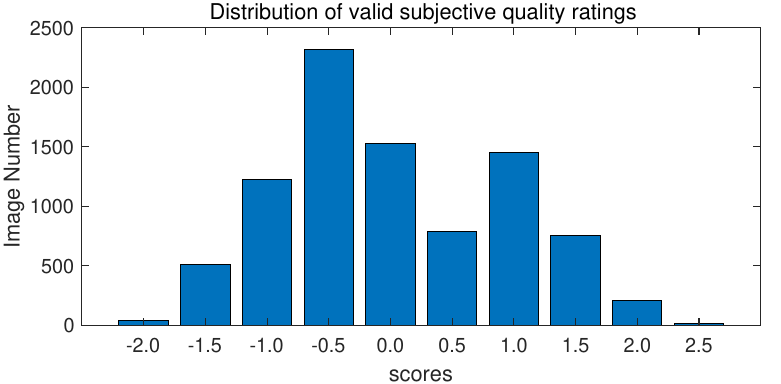}
  \vspace{-3.0mm}
  \caption{Distribution of subjective quality ratings.}
  \vspace{-5.0mm}
  \label{rating_distribution}
\end{figure}

\vspace{-3.5mm}
\subsection{Data Processing and Analysis} \label{section:sec3.3}
We follow the suggestions given in \cite{series2012methodology,duan2022confusing,duan2022augmented} to conduct the outlier detection and subject rejection.
Specifically, we first calculate the kurtosis score of the raw subjective quality ratings to judge if it belongs to a Gaussian or non-Gaussian category.
For Gaussian distribution, the original rating is considered an outlier if it exceeds two standard deviations from the average rating of the image \cite{sun2024visual,zhu2023audio}.
For non-Gaussian distributions, outliers are determined when they fall outside the $\sqrt{20}$ criterion relative to the average rating of the image \cite{duan2022confusing}.
Then we reject subjects whose data offset exceeds 5\%.
As a result, 1 subject is screened. From the ratings submitted by the remaining valid subjects, about 2.38\% of the total subjective evaluations are flagged as outliers and subsequently eliminated. Then we compute the average rating for each image to derive the final mean opinion score (MOS) as follows.
\begin{equation}
  z_{ij}=\frac{m_{ij}-\mu_{i}}{\sigma_{i}},\quad z_{ij}'=\frac{100(z_{ij}+3)}{6},
\end{equation}
\vspace{-10pt}
\begin{equation}
  MOS_{j}=\frac{1}{N}\sum_{i=1}^{N}z_{ij}',
\vspace{-3pt}
\end{equation}
where $m_{ij}$ is the raw rating given by the $i$-th subject to the $j$-th image, $\mu_{i}$ is the mean rating given by subject $i$, $\sigma_{i}$ is the standard deviation, and $N$ is the total number of subjects.
The distribution of the valid quality ratings is shown in Fig. \ref{rating_distribution}.

\begin{figure*}[t]
  \centering
  \includegraphics[width=0.95\textwidth]{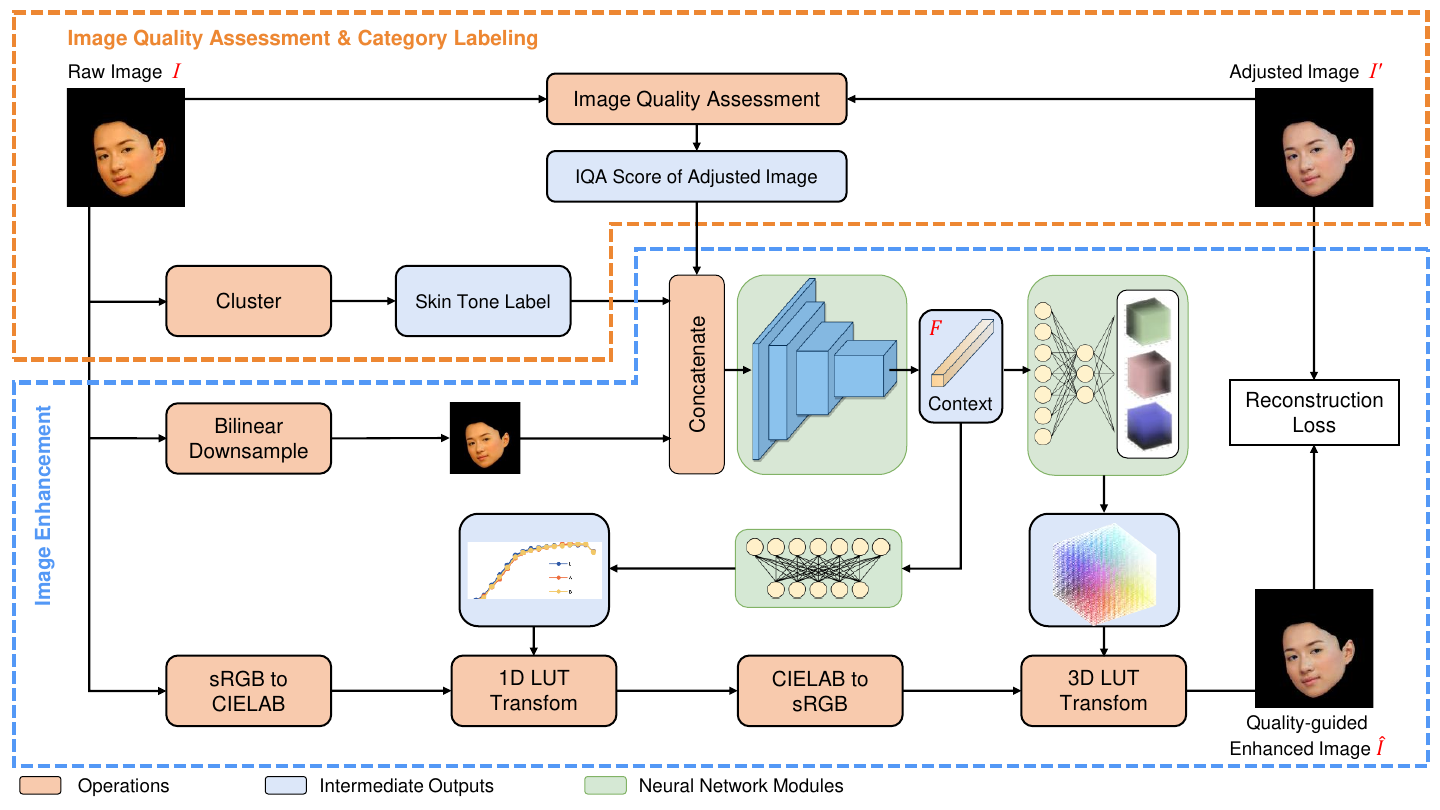}
  \vspace{-3.2mm}
  \caption{An overview of the framework of our method. The image quality assessment and category labeling modules are shown in the \textcolor{NewOrange}{orange} dotted box. The image enhancement modules are shown in the \textcolor{NewBlue}{blue} dotted box.}
  \vspace{-5.0mm}
  \label{framework}
\end{figure*}

\vspace{-2.5mm}
\section{Quality-guided Image Enhancement Method}
In this section, we illustrate the proposed quality-guided image enhancement method in detail.
Fig. \ref{framework} demonstrates the overview of the proposed method for skin tone enhancement based on a learnable 3D LUT.
The details for each module are introduced as follows.

\vspace{-3.5mm}
\subsection{Overall Pipeline}

The upper part in the \textcolor{NewOrange}{orange} dotted box of Fig. \ref{framework} represents the IQA and category labeling module, which generates the relative skin tone quality rating to guide the enhancement procedure and produces the skin tone label to improve the adaptability of the module. 
These two values are concatenated with the down-sampled raw image to serve as the input to the CNN network. 
The lower part in the \textcolor{NewBlue}{blue} dotted box of Fig. \ref{framework} constitutes the image enhancement module which is built based on a lightweight image enhancement 3D LUT model \cite{zeng2020learning,yang2022seplut}.
Similarly, we employ a cascade of three 1D LUTs and one 3D LUT to construct the network. 
Specifically, the CNN backbone receives the concatenated input image and guided information, and outputs a feature vector containing rich context information \cite{zhou2022pyramid,zhou2023quality}. 
The image context is used to guide the generation of three 1D LUTs and a 3D LUT in an image-adaptive fashion \cite{zeng2020learning,yang2022seplut}.
For the adjustment process, we first convert the image from the sRGB color space to the CIELAB color space before performing the 1D LUT operations since the skin tone adjustment in our STIQAD is mainly performed on the CIELAB space. 
Next, the three channels ($L$, $A$, and $B$) are independently operated by three 1D LUTs, resulting in an initial adjustment of brightness and color.
Subsequently, the image is reverted to the RGB color space to look up the corresponding color in the RGB color space according to the fused 3D LUT to achieve enhancement.
The reconstruction loss is finally calculated between the enhanced image and the adjusted image.

\vspace{-3.5mm}
\subsection{Quality-guided Strategy} \label{section:sec4.2}
In this paper, we introduce a novel image enhancement strategy that relies on quality scores as a guiding principle. 
The main idea is to utilize the raw image and the corresponding quality score of the adjusted image as the input for the enhancement network, and subsequently optimize the network to make the enhanced output image as close as the adjusted image. 
The IQA module plays a role in providing relative quality scores. 
In this work, the quality rating was derived from the subjective experiment in Section \ref{section:sec3}. 
However, it can be substituted by any other IQA method capable of providing an accurate quality rating, as shown in Section \ref{section:sec5.4}. 
By leveraging these quality rating data, the image enhancement module learns the distribution characteristics of images with various quality scores, thereby achieving quality-guided image enhancement and being able to adjust an image continuously.
It is worth mentioning that although we choose the 3D LUT-related image enhancement method in our experiment, our proposed approach has the potential to be applied to other image enhancement algorithms with differentiable properties as discussed in Section \ref{section:sec5.4}.

\vspace{-3.5mm}
\subsection{Skin Tone Label}
When adjusting the skin tone of a portrait photograph, various original skin tones may exhibit distinct adjustment preference directions. 
The skin tone adjustment for facial regions necessitates more stringent criteria compared to other image adjustment requirements. 
Consequently, even if the existing 3D LUT model possesses image adaptability, it is challenging to directly learn these diverse enhancement directions. 
However, for images with similar skin tones, the directions of quality improvement or degradation are more alike. 
To address this issue, we use clustering methods to determine category centers which can be used to classify the collected images into several tone categories. 
During training or inference, we first determine the tone category of an image, and then concatenate the classification tone label to the image and the quality score as the input to the network. 
The weights of 1D LUTs and 3D LUTs are acquired through training on the combination of the original images, quality scores and skin tone labels.

\begin{figure}[t]
  \centering
  \includegraphics[width=0.95\linewidth]{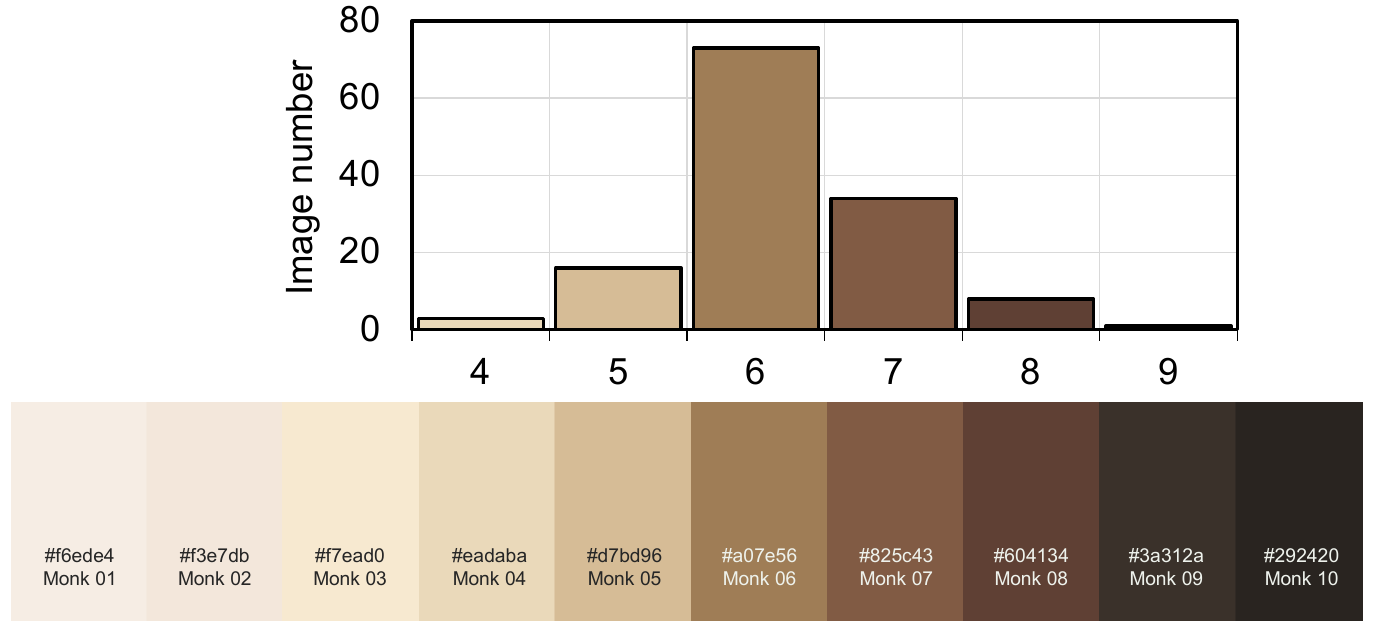}
  \vspace{-4.0mm}
  \caption{Skin tone image classification results based on Google monk skin tone.}
  \vspace{-5.0mm}
  \label{google_center}
\end{figure}

\begin{figure}[t]
  \centering
  \includegraphics[width=0.65\linewidth]{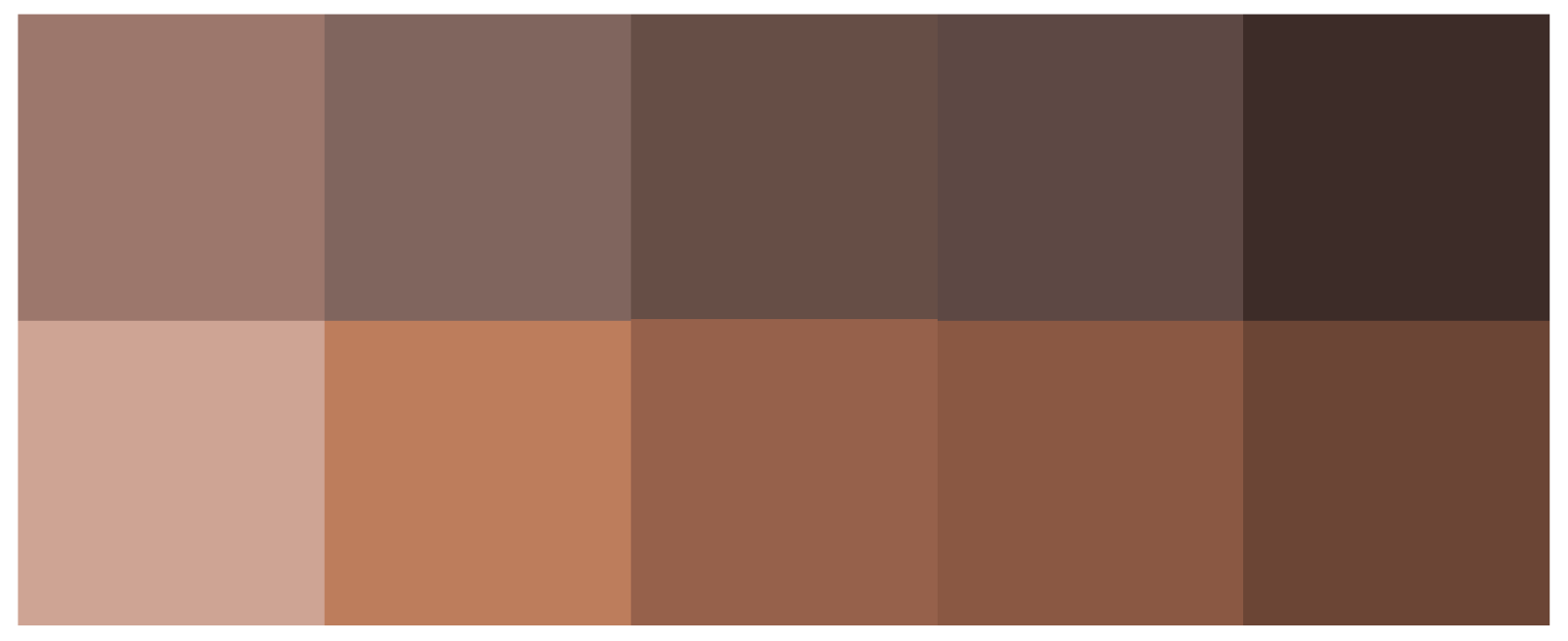}
  \vspace{-3.0mm}
  \caption{Skin tone centers generated by our proposed clustering-based method.}
  \vspace{-6.0mm}
  \label{center}
\end{figure}

Specifically, the generation of skin tone labels needs two steps, a center definition step and a label calculation step. 
The first step requires calculating a set of reliable clustering centers by using a large number of images containing the facial region with various skin tones. 
The second step involves assigning the existing images to the nearest center.
In the first step, we first consider directly using the Monk Skin Tone Scale \cite{google} provided by Google as the clustering center. 
We use a face parsing model to segment the facial region and calculate the average color value of each face from a dataset of 135 faces with various skin tones. 
Then, we use the L2 distance to classify each image based on its distance of the ten centers of the Google Monk Skin Tone Scale. The classification results are shown in Fig. \ref{google_center}.
The figure shows that most of the 135 images are distributed in categories 5-8, with a very small distribution in categories 4 and 9, and no distribution in the rest of the categories. 
In fact, the 135 portrait photos we selected include a wide range of skin tone images. 
Considering that Google's centers are based on the classification of human skin color in an ideal situation, while our main task is to process the skin of the human face in photos, actual skin tones will be influenced by various lighting conditions during photography. 
Therefore, we propose to extract skin colors from photos and cluster them to obtain cluster centers discussed as follows.

\begin{figure}[t]
  \centering
  \includegraphics[width=1.0\linewidth]{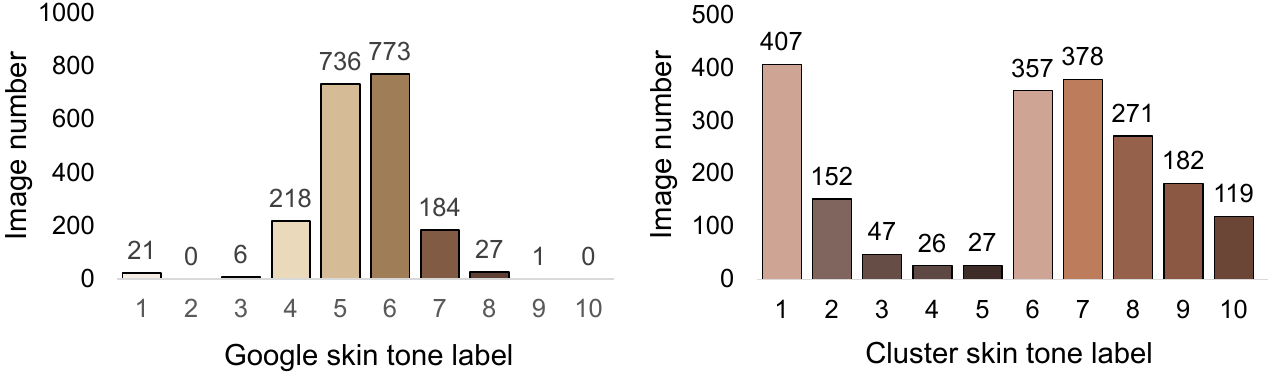}
  \vspace{-7.0mm}
  \caption{Skin tone image classification results on the Helen dataset.}
  \vspace{-7.0mm}
  \label{Helen}
\end{figure}

To obtain category centers that better fit the skin tones of faces from the photography, we select 1673 images from the Lapa dataset \cite{liu2020new}, 334 clear face images from the Fivek dataset \cite{bychkovsky2011learning}, and 1024 portraits collected by camera devices. 
The dataset contains a variety of skin tones with relatively uniform distribution, and all images are used face parsing method to leave the skin tone region for calculating the category centers. 
By using K-means clustering method in the perceptually uniform CIELab space on the above images, we obtain ten clustered category centers, as shown in Fig. \ref{center}.
To verify the clustering performance of the obtained centers, we select another commonly used Helen dataset \cite{le2012interactive} to test the classification performance. The test results are shown in Fig. \ref{Helen}.
It can be observed that compared to Google's skin tone centers, our proposed clustering method has a more scattered distribution on the Helen dataset.
The Silhouette Coefficient obtained by using our clustered center for classification is 0.18, while using Google skin tone as the classification center on this dataset is 0.13, which further manifests the effectiveness of our method. 

%



\vspace{-4.5mm}
\subsection{Image Enhancement Module} \label{section:sec4.3}
We mainly adopt the learnable 3D LUT method as the image enhancement module in this work, which mainly includes a CNN backbone part for global image context analysis and a learnable look up table part for image adjustment.

{\bfseries\itshape CNN backbone for global image context analysis.} A lightweight CNN network is adopted as the backbone to receive 5-layer image information including 3 layers for the down-sampled image, a layer for the category label and a layer for the quality score. 
The score and label are replicated to match the dimensions of the down-scaled image. 
The CNN backbone contains 5 convolutional layers with instance Norm between them.
The output context feature vector $F$ is then fed into two multi layer perceptron (MLP) networks for LUT generation.

{\bfseries\itshape Learnable lookup table.} 
3D LUT is utilized for the comprehensive pixel mapping of an image. A 3D LUT based on the RGB space aims to minimize the RGB distance between the original image and the adjusted image. 
However, due to the poor perceptual uniformity of the RGB color space and the adjustment of the images in our STIQAD is mainly conducted based on the LAB space, we propose to use three 1D LUTs before the 3D LUT to pre-process the three CIELAB channels individually \cite{yang2022seplut}. CIELAB is a color space with a certain level of perceptual uniformity introduced by the International Commission on Illumination (CIE) in 1976 \cite{robertson1977cie}, which represents brightness information, red-green, and yellow-blue components with three axes, $L$, $a$, and $b$. Xu \textit{et al.} \cite{xu2021testing} demonstrated the perceptual consistency advantage of CIELAB compared to RGB. By employing three 1D LUTs for the CIELAB channels, the brightness, chroma, and hue of the image can be pre-adjusted to more closely resemble the target image's state.

With the acquired context feature vector $F$, the 1D LUT transform performed on the image $I$ can be formulated as: 
\vspace{-1.0mm}
\begin{equation}
    \{T^l_{1D},T^a_{1D},T^b_{1D}\} = g_{1D}(F),
    \vspace{-0.5mm}
\end{equation}
where $T^l_{1D},T^a_{1D},T^b_{1D}$ denotes the obtained three 1D LUTs for $L$, $A$, $B$ channels, respectively, $g_{1D}$ is the 1D LUT generator module that takes the feature vector $F$ as input.
The 3D LUT transform can be formulated as:
\vspace{-1.0mm}
\begin{equation}
    T_{3D} = g_{3D}(F),
    \vspace{-1.0mm}
\end{equation}
where $T_{3D}$ denotes the obtained 3D LUT, $g_{3D}$ is the 3D LUT generator module that takes the feature vector $F$ as input.
Then the image enhancement process can be formulated as:
\vspace{-1.0mm}
\begin{equation}
    \hat{I} = T_{3D}(\text{LAB2RGB}(T_{1D}(\text{RGB2LAB}(I)))).
    \vspace{-1.0mm}
\end{equation}

The overall loss function can be formulated as:
\vspace{-1.0mm}
\begin{equation}
    \begin{split}
    \mathcal{L} = \lambda_{1}\mathcal{L}_{r} + \lambda_{2}\mathcal{L}_{s} +\lambda_{3}\mathcal{L}_{m},
    \end{split}
    \vspace{-5.0mm}
\end{equation}
where $\lambda_{1}$, $\lambda_{2}$, $\lambda_{3}$ are hyperparameters to balance the loss items, which are set to 1, $1\times10^{-4}$, 10, respectively.

\begin{figure*}[t]
  \includegraphics[width=0.9\linewidth]{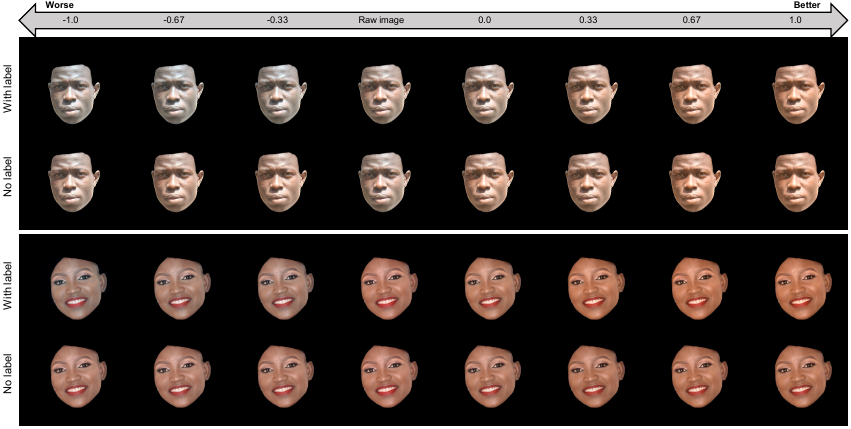}
  \centering
  \vspace{-3.0mm}
  \caption{Quality-guided image enhancement on portrait photos with darker skin tones. Above the images are the guiding quality scores, ranging from -1 to 1. ``With label'': the results of the model with skin tone label module. ``No label'': the results of the model without skin tone label module.}
  \vspace{-4.5mm}
  \label{daker_enhance}
\end{figure*}

\begin{figure*}[t]
  \includegraphics[width=0.9\linewidth]{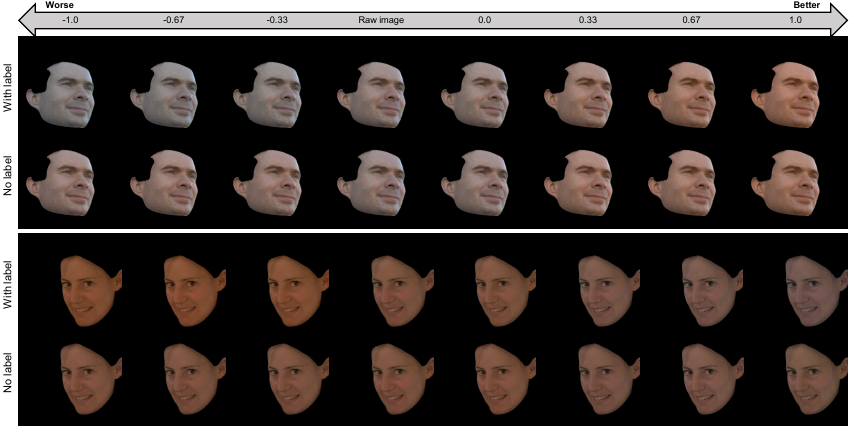}
  \centering
  \vspace{-3.5mm}
  \caption{Quality-guided image enhancement on portrait photos with lighter skin tones. Above the images are the guiding quality scores, ranging from -1 to 1. ``With label'': the results of the model with skin tone label module. ``No label'': the results of the model without skin tone label module.}
  \vspace{-1.5mm}
  \label{lighter_enhance}
\end{figure*}

\begin{figure*}[!t]
  \includegraphics[width=0.96\linewidth]{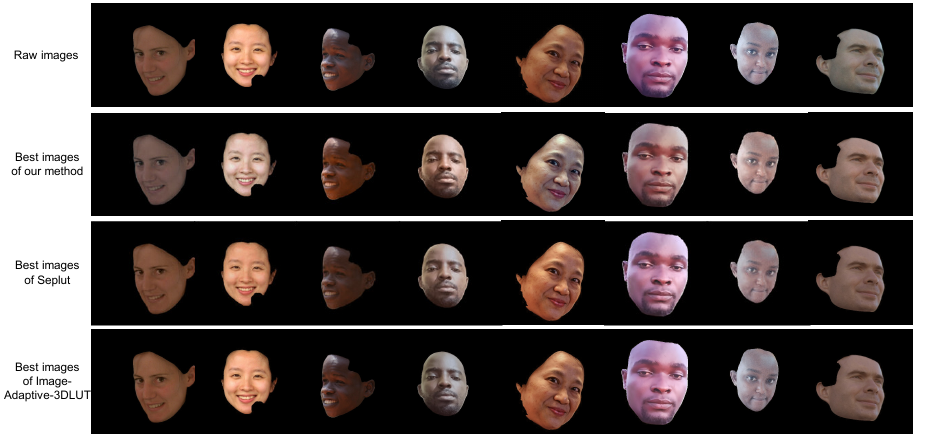}
  \centering
  \vspace{-3.0mm}
  \caption{Best images obtained from our method, Seplut and Image-Adaptive-3DLUT}
  \vspace{-5.1mm}
  \label{best_images}
\end{figure*}

\vspace{-1.5mm}
\section{Experimental Validation}
\subsection{Enhancement for Skin Tone} 
\subsubsection{Experimental Data and Setup} \label{section:sec4.1.1}
We first conduct a quality-guided experiment on STIQAD
for facial skin tone adjustment. 
Following the procedure outlined in Section \ref{section:sec3.3}, we acquire a total of 1105 pairs of raw images and adjusted images with the corresponding quality ratings. 
Specifically, we first treat the 85 original images as both input images and adjusted images, and randomly assign quality ratings with absolute values within 0.1. 
This approach assists our output in closely approximating the raw image when the guidance score is 0. 
We divide the 1190 pairs of images into 1105 pairs for training and 85 pairs for testing. 
Additionally, we select 50 new test images to further extensively examine the subjective visual effects of enhanced images guided by various scores. 
All quality ratings are normalized to [-1, 1] before being concatenated to the image.

We follow the hyper-parameter settings of Zeng \textit{et al.} \cite{zeng2020learning}, utilizing the standard Adam optimizer with the mini-batch size of 1. 
All models are trained for 400 epochs with a fixed learning rate of 1$e^{-4}$ on an NVIDIA Tesla 2080ti GPU. 

\subsubsection{Experimental Results and Analysis}
Fig. \ref{daker_enhance} and Fig. \ref{lighter_enhance} demonstrates the adjusted images guided by various quality scores.
The input guiding score can vary continuously from -1.0 to 1.0, and we display the visualization results with the guiding score of -1.0, -0.67, -0.33, 0.0, 0.33, 0.67 and 1.0.
It can be observed that with the help of our proposed framework, we can enhance images continuously and controllably based on various quality scores.
Our experimental results also show that the proposed model can adaptively identify the adjustment directions for portrait photos with different original skin tones. 
More visualization results can be found in the supplementary materials.

We further conduct additional experiments to evaluate the effectiveness of our proposed framework by comparing with  two state-of-the-art image enhancement methods including Image-Adaptive-3DLUT \cite{zeng2020learning} and Seplut \cite{yang2022seplut}.
Since these existing models \cite{zeng2020learning,yang2022adaint,yang2022seplut} lack the ability of performing controllable and continuous image enhancement, we make single-level comparisons across the best results of using Image-Adaptive-3DLUT, Seplut and our framework.
It should be noted that these existing models can only learn the one-to-one image mapping, thus each raw image can only correspond to one adjusted image.
Therefore, to obtain the best images from Seplut and Image-Adaptive-3DLUT for fair comparison, we first pick up the image with the highest MOS from the adjusted images in our STIQAD, then regard these images as the ground truths to train these existing image enhancement models.
All 85 pairs of images from STIQAD are used for training and 20 images which are previously unseen during the training stages are utilized for testing.
For comparison, the best images in our framework are obtained by setting the input quality score to 1.
The experimental results demonstrated in Fig. \ref{best_images} qualitatively show that our proposed method achieves better enhanced results in visualization compared to Image-Adaptive-3DLUT and Seplut.
To quantitatively evaluate the effectiveness of our method, we conduct a subjective experiment to compare the results generated by our method and the other two methods \cite{zeng2020learning,yang2022seplut}, respectively.
Specifically, during the subjective experiment, subjects are shown with two images, one of which is the best image generated by our proposed method and the other is the best image generated by an existing LUT method (\cite{zeng2020learning} or \cite{yang2022seplut}).
Subjects are asked to evaluate the subjective perceptual quality of the two images and choose the better image or choose that the two images have the similar perceptual quality.
A total of 15 subjects are recruited to give their opinions.
According to our survey results shown in Table \ref{tab:evaluation}, when comparing our proposed method to Image-Adaptive-3DLUT, a majority of opinions (58.7\%) indicate that our proposed method generates better enhancement results.
In contrast, only 13.0\% of opinions prefer Image-Adaptive-3DLUT in terms of image quality.
28.3\% of the opinions indicate that the subjective quality is perceived similar.
For the comparison with Seplut, 48.4\% of opinions prefer our proposed method, while 20.8\% prefer Seplut.

\begin{table}
   \centering
   \caption{Evaluation results of comparing the best enhanced images generated by our method and the other two methods.}
   \vspace{-2.0mm}
   \resizebox{\linewidth}{!}{
   \begin{tabular}{c|ccc}
       \toprule
       Options & Our Method & Image-Adaptive-3DLUT & Similar Quality \\ \cmidrule(lr){1-4}
       Percentage (\%) & \bfseries{58.7} & 13.0 & 28.3 \\ \midrule
       Options & Our Method & Seplut & Similar Quality \\ \cmidrule(lr){1-4}
       Percentage (\%) & \bfseries{48.4} & 20.8 & 30.8 \\ \bottomrule
   \end{tabular}
   }
   \label{tab:evaluation}
   \vspace{-5.6mm}
\end{table}

\begin{figure*}[t]
  \centering
  \includegraphics[width=0.88\linewidth]{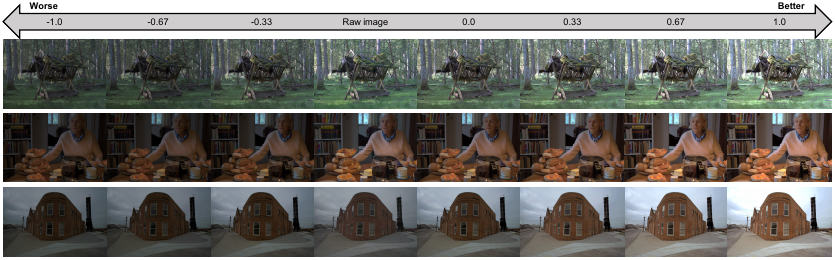}
  \vspace{-3.0mm}
  \caption{Quality-guided image enhancement on natural images. Above the images are the guiding quality scores, ranging from -1 to 1.}
  \vspace{-3.0mm}
  \label{natural}
\end{figure*}

\begin{figure*}[t]
  \centering
  \includegraphics[width=0.91\linewidth]{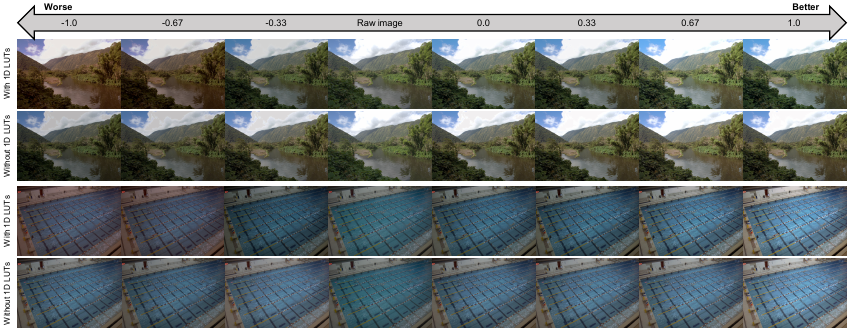}
  \vspace{-3.0mm}
  \caption{Comparisons between using 1D LUTs and not using 1D LUTs.}
  \vspace{-5.0mm}
  \label{1dlut}
\end{figure*}

\vspace{-4.5mm}
\subsection{Enhancement for Natural Images}
As aforementioned, we have conducted a study focusing on facial skin tone adjustment, which verifies the effectiveness of our method in adjusting color. 
In this section, we further carry out a natural image enhancement experiment based on our proposed quality-guided enhancement method to demonstrate that our method can be applied not only to facial image enhancement, but also to natural image enhancement, and our method can perform person-specific enhancement based on a small number of images. 
The specific experimental steps are described below.

\subsubsection{Experimental Data and Setup}
We choose the first 10 images from the FiveK dataset \cite{bychkovsky2011learning} for the training process. 
Similar to the data adjustment method used for portraits, we also randomly adjust the ten natural images and employ the same screening method. 
Each image generates 13 adjustments, which include various levels of subjectively perceived improvements and degradations. 
The first 10 images from the FiveK dataset and their corresponding adjusted images are only used for training.
We use the 11-th to 35-th images for testing. All testing images are not seen during the training stage in our experiments.
It is worth noting that, in the skin tone adjustment experiment, we aim to eliminate the influence of brightness information on subjective quality perception, so we maintain consistency in image brightness.
However, for the natural image adjustments, we no longer restrict the consistency of brightness information but modify the three values of the CIELAB color space.
In natural image enhancement, we only use quality labels and remove the skin tone label part. Unlike the enhancement for skin tone, the enhancement for natural images tends to be an overall adjustment of the distribution. Therefore it is not easy to distinguish different adjustment tendencies in natural images. If it is possible to separate the different adjustment tendencies, we can also add labels to help improve adaptability.

\begin{figure*}[t]
  \centering
  \includegraphics[width=0.9\linewidth]{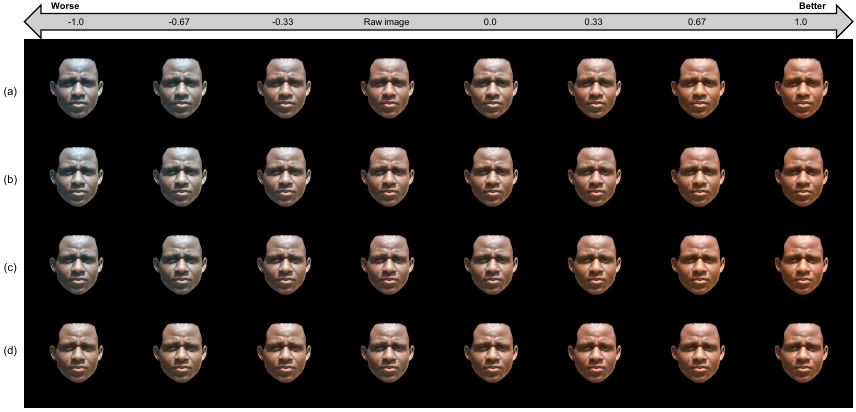}
  \vspace{-4.0mm}
  \caption{Comparisons between different normalization strategies. (a) Scores are normalized to the range of -1 to 1, labels keep the initial range of 1 to 10. (b) Scores are normalized to the range of 0 to 1, labels keep the initial range of 1 to 10. (c) Scores are normalized to the range of -5 to 5, labels keep the initial range of 1 to 10. (d) Scores are normalized to the range of -1 to 1, labels are normalized to the range of 0 to 1. The numbers above the images are the guiding quality scores, ranging from -1 to 1.}
  \vspace{-5.5mm}
  \label{norm}
\end{figure*}

To validate that our method can achieve individual-specific enhancement rather than relying on multiple participants to obtain more reliable and stable scores, we only collect individual quality score data.
All settings remain the same as in the skin tone test. The subject observes the original image and the adjusted natural image, assigning a rating between \mbox{-2.5} and 2.5 to the adjusted image.
Finally, the single score result obtained is normalized from the range of [-2.5, 2.5] to [-1, 1], without calculating the MOS.

\subsubsection{Network Framework}
Compared to the skin tone adjustment experiment, the natural image experiment reduces the input of the skin tone label. 
The network directly concatenates the three color channels of the raw image and the quality score of the adjusted image into four channels, which are then fed into the model.
After applying three separate 1D LUT operations in the CIELAB color space and the 3D LUT operations to the raw image, the loss between the adjusted image and the enhanced image is calculated.
The remaining network structure and implementation details are consistent with those described in Section \ref{section:sec4.3}.

\begin{figure}[t]
  \centering
  \includegraphics[width=\linewidth]{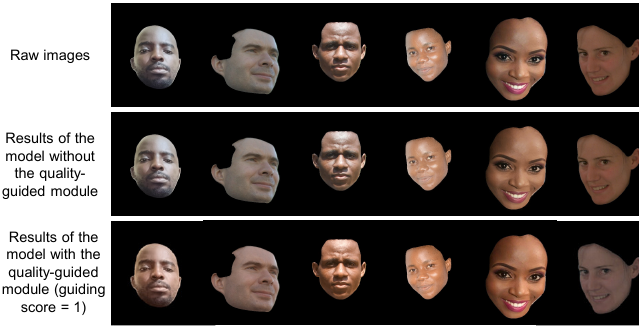}
  \vspace{-8.0mm}
  \caption{Comparisons between the results of the model with or without the quality-guided module.}
  \vspace{-6.5mm}
  \label{without_score}
\end{figure}

\subsubsection{Experimental Results and Analysis} 
The outcomes of the quality-guided enhancement for natural images are presented in Fig. \ref{natural}.
The results demonstrate that even with a single individual rating and a limited number of adjusted images, our method can still achieve image enhancement guided by scores.
This also implies that in practical application scenarios, we can obtain quality ratings from the users by presenting them with paired images and enhance images according to their preferences, which demonstrates the personalized enhancement ability of our proposed framework.
More visualization results are shown in the supplementary materials.

\vspace{-4.5mm}
\subsection{Ablation Experiment}
\vspace{-0.5mm}
\subsubsection{The Use of Skin Tone Label}
In order to assess the impact of skin tone labels on the experimental results, we conduct an experiment without the skin tone label and compare the outcomes.
As PSNR cannot serve as a criterion for subjective perception, we only present the enhanced image for qualitatively comparison in this paper as demonstrated in Fig. \ref{daker_enhance} and Fig. \ref{lighter_enhance}.
When the skin tone label is not included, the enhancement results of the model exhibit a significant reduction in discrimination with various quality scores.
Moreover, without skin tone labels, the adjustment directions of some images may be wrong. Consequently, including skin tone labels in our framework for skin tone enhancement is essential.
More ablation results can be found in the supplementary materials.

\subsubsection{The Use of 1D LUTs}
We also study the impact of the use of 1D LUTs.
As demonstrated in Fig. \ref{1dlut}, applying 1D LUTs to the L, A and B channels in the CIELAB colorspace can improve the color and brightness distinguishing degree.

\subsubsection{The use of IQA score}
We further conduct an experiment without the IQA score module on our database. 
The results shown in Fig. \ref{without_score} demonstrate that the model cannot work without the IQA score module on our database, since our database contains different adjusted images for one raw image, which is beyond the capability of the basic 3D LUT method.
And it should be noted that without the IQA score, our framework cannot achieve controllable and continuous enhancement.

\subsubsection{Normalization of Scores and Labels}
In this section, we further explore the effect of various normalization strategies of scores and labels for the final performance.
Generally, we normalize the MOSs to the range of [-1, 1] and the skin tone labels to 10 integers in the range of [1, 10].
Here, we explore three variants, which includes normalizing the scores to the range of [0, 1] and [-5, 5], and normalizing the skin tone labels to [0, 1], respectively.
The results are illustrated in Fig. \ref{norm}.
It qualitatively manifests that the alterations of score normalization do not significantly affect the model training.
However, compressing the skin tone label range to 0-1 results in a decrease in the differentiation between the adjusted images across various quality grades.

\begin{figure*}[!t]
  \centering
  \includegraphics[width=0.86\linewidth]{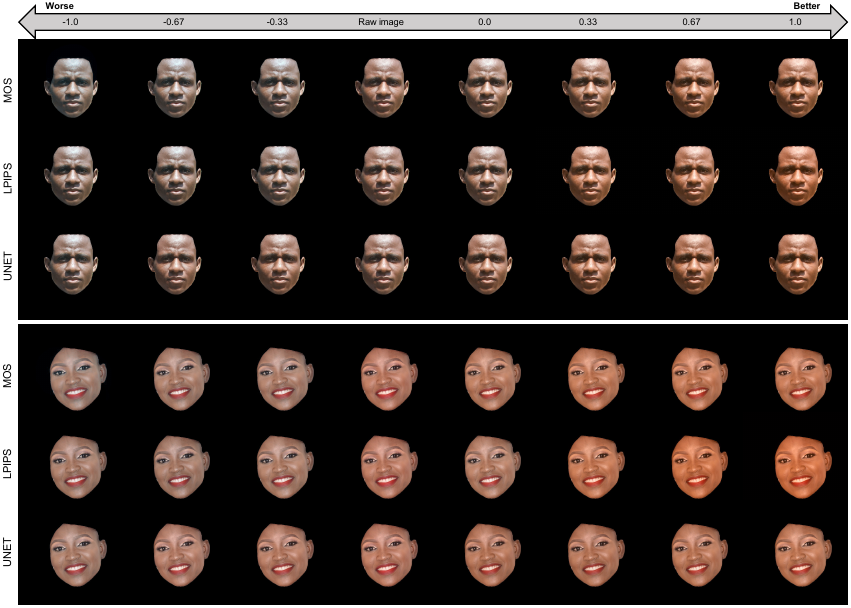}
  \vspace{-3.5mm}
  \caption{``MOS'': image enhancement guided by subjective MOS, enhanced by 3D LUT-related method. ``LPIPS'': image enhancement guided by LPIPS value, enhanced by 3D LUT-related method. ``UNET'': image enhancement guided by subjective MOS, enhanced by UNet.}
  \vspace{-5.5mm}
  \label{LPIPS}
\end{figure*}

\begin{figure}[t]
  \centering
  \includegraphics[width=0.94\linewidth]{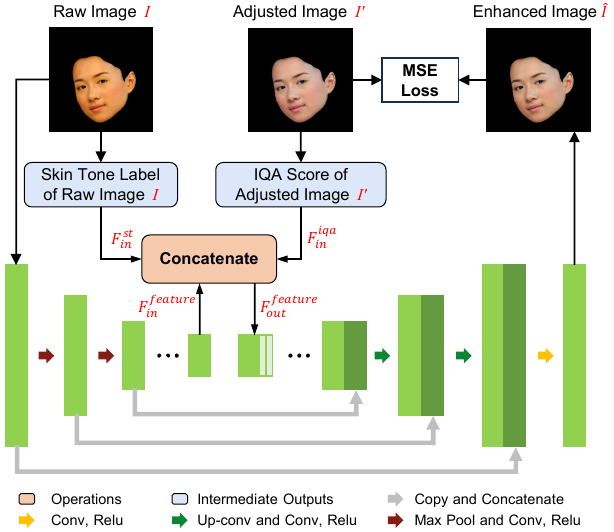}
  \vspace{-4.0mm}
  \caption{An overview of the framework using UNet.}
  \vspace{-6.5mm}
  \label{unet}
\end{figure}

\vspace{-3.0mm}
\subsection{Generalization Capability of the Framework} \label{section:sec5.4}
\vspace{-1.5mm}
As aforementioned, we use the subjective IQA scores to control the enhancement process of the 3D LUT enhancement model.
However, as mentioned in Section \ref{section:sec4.2}, both the IQA module and the image enhancement module can be substituted by other method. 
To further validate that the proposed framework has generalization abilities on other guided methods or other enhancement methods, we conduct two additional experiments, which include using LPIPS \cite{zhang2018unreasonable} as the IQA part of the framework and using UNet \cite{ronneberger2015u} as the image enhancement part.

\subsubsection{IQA Module Substitution}
We first apply LPIPS \cite{zhang2018unreasonable} as the IQA module of the framework to prove that the quality rating can be replaced by other IQA method. Since objective IQA scores are more easily available than subjective IQA scores, utilizing objective scores can greatly improve the generalization ability of our method.
It should be noted that most of the existing full-reference (FR) image quality assessment (IQA) models are difference/similarity evaluation metrics \cite{zhang2018unreasonable,wang2023measuring,chen2023learning}, thus lacking the ability to determine the perceived quality of an image is better or worse compared to another image.
It is hard to directly use these FR-IQA methods in our framework since they cannot determine the changing direction (better/worse). 
Therefore, to use LPIPS to guide image enhancement in our work, we train a simple classifier based on the VGG network to predict whether the adjusted image results in better or worse in terms of the perceptual quality compared to the raw image.

Briefly, we input raw and adjusted portrait image pairs into the VGG net and output binary results (better/worse).
If the classifier determined that the perceptual quality is worse, we multiply the LPIPS value of the two images by -1. Conversely, if the quality is better, we multiply the LPIPS value by 1. Then we guide the image enhancement training process by these LPIPS scores with better or worse labels.

Fig. \ref{LPIPS} demonstrates the comparison results of using MOS and LPIPS scores for quality guidance, respectively.
The experimental results show that LPIPS can also successfully guide the image adjustment continuously.
However, it is deserved to be noted that the enhancement results of the model trained using LPIPS are not as satisfactory as the model trained using subjective MOS.
Specifically, as the input score increased, a subset of the face images exhibit a gradually darker orange color, which do not align with the preferences of most individuals.
Moreover, the continuous adjustment guided by LPIPS is uneven compared to MOS-guided. The rate of change guided by LPIPS is non-uniform.
This suggests that better IQA models for portrait photography are needed. 
We will study this problem in our future work.

\subsubsection{Image Enhancement Module Substitution}
We also demonstrate that the proposed method has the potential to be applied to other image enhancement algorithms.
In this experiment, we conduct the quality-guided image enhancement experiment based on UNet instead of 3D LUT-related method.
Fig. \ref{unet} shows the overview of the proposed framework for skin tone enhancement based on UNet.
The skin tone label and the IQA score are concatenated with the deepest feature map of UNet.
Then the MSE loss between the adjusted image and the enhanced image is calculated.
We keep the training and testing data as the same with that in Section \ref{section:sec4.1.1}.
Some visualization results are presented in Fig. \ref{LPIPS}.
The effect of using UNet as the backbone is not as obvious as using 3D LUT, however, the continuous trend of image quality can still be clearly observed.
More visualization results to demonstrate the generalization capability of our framework are shown in the supplementary materials.

\begin{figure*}[t]
  \centering
  \includegraphics[width=0.89\linewidth]{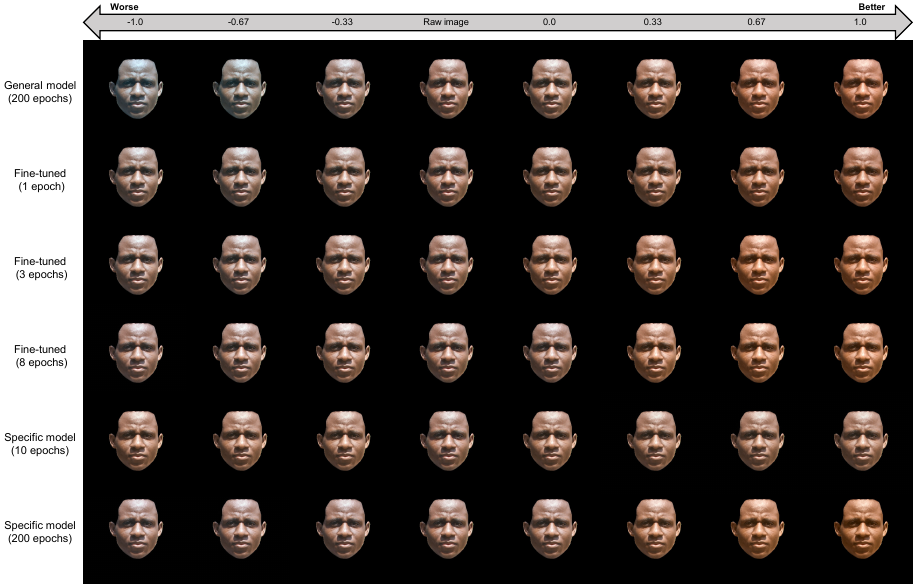}
  \vspace{-3.5mm}
  \caption{
  Comparisons between different models. ``General model'' indicates the model trained with the MOSs without one specific user's ratings. ``Specific model'' represents the model trained using the scores of one specific user. ``Fine-tuned'' means the model fine-tuned from the general model using the scores of the specific user. 
  }
  \vspace{-6.0mm}
  \label{tune}
\end{figure*}

\vspace{-3.5mm}
\subsection{Fine-tuning the General Model with Specific Opinions} \label{section:sec5.5}
We conduct an experiment to fine-tune the general model with the opinions of a specific user.
We first randomly select the scores of a single individual from the collective set of scores obtained in our subjective experiment.
These scores are chosen as the specific opinion scores of the user.
The scores of all other participants are calculated as mean opinion scores (MOS) to act as general scores.
Then we use these two kinds of scores (specific opinion scores \& mean opinion scores) to train two models separately for 200 epochs, the results of which are shown in the first and last rows of Fig. \ref{tune}.
To validate the superiority of our pre-trained general model, we fine-tune the trained general model using the specific opinion scores of the single user.
The results show that fine-tuning the model initialized from the pre-trained general model with only 10 epochs can achieve the effect of the model trained from scratch with about 200 epochs, which also demonstrates that our proposed method is adaptable.

\vspace{-4.5mm}
\subsection{Experiment with Multiple Enhancement Rounds}
\begin{figure*}[t]
  \centering
  \includegraphics[width=0.85\linewidth]{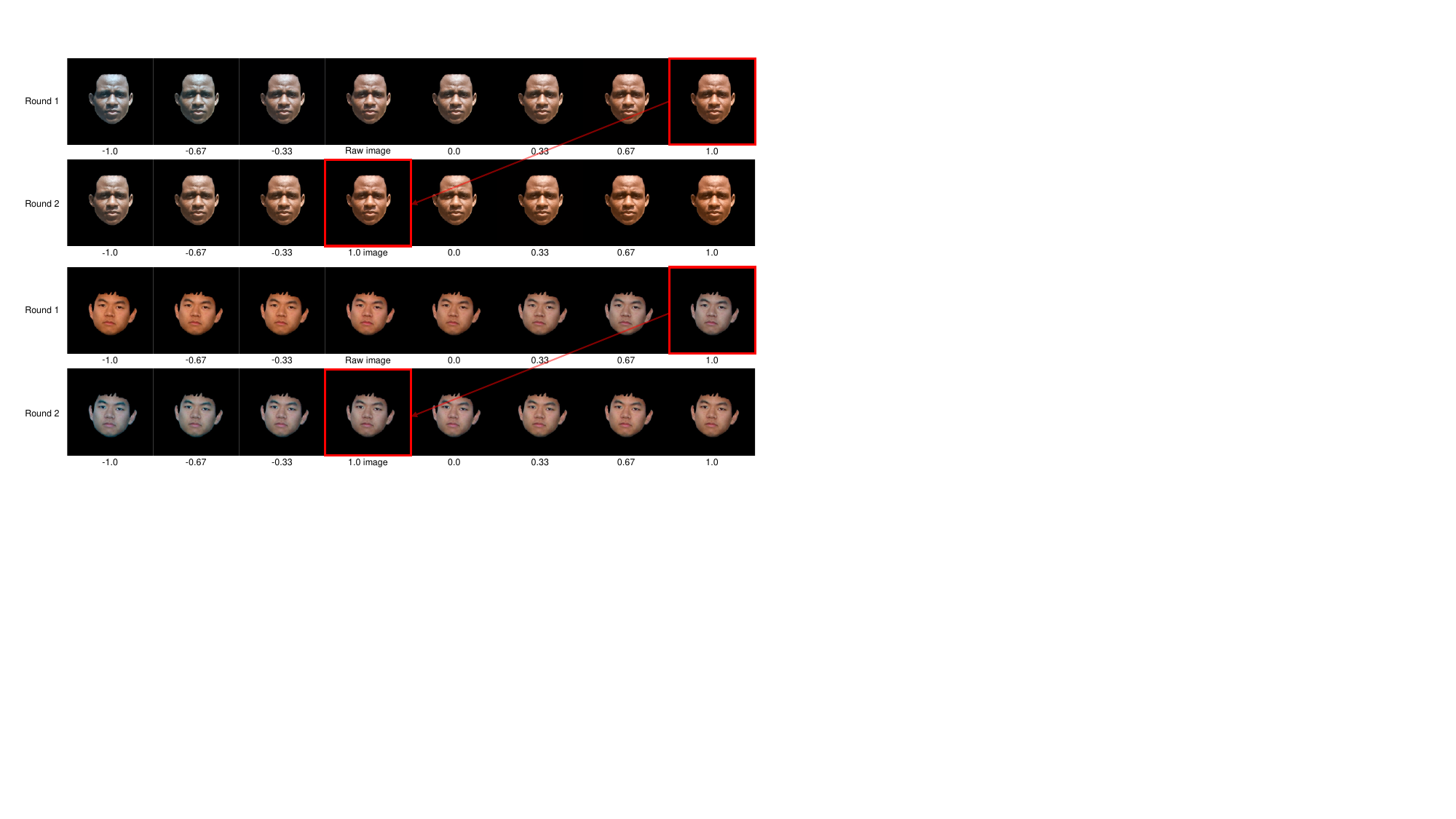}
  \vspace{-3.5mm}
  \caption{Sequential dual-round quality-guided image enhancement. The images to be enhanced in the second round are from the images with the highest score in the first round.}
  \vspace{-4.5mm}
  \label{round}
\end{figure*}

\vspace{-0.5mm}
Our proposed framework can adjust images continuously and controllably with given quality scores, however, in some cases, we may want to further enhance the images beyond the score range.
In this experiment, we further investigate the abilities of the proposed framework on continuously enhancing an image based on the adjusted image with the highest quality score.
we first set the input quality score to 1 to generate the highest score-enhanced images, then use these images as the initial images and conduct the second-round experiment based on the proposed framework. 
The results are presented in Fig. \ref{round}.

For the images in the first two rows of Fig. \ref{round}, the enhancement trend for the second round is largely consistent with the first round, but the images can be further enhanced slightly.
In contrast, for the images in the last two rows of Fig. \ref{round}. In the first round, the original portrait was overly yellow and this situation is mitigated following enhancement.
The second round of enhancements further adjusted the coloration, making the subject appear more ruddy.
The trend in the second round of enhancement diverges from the first round and the second-round enhancement can further correct the slightly over-enhancement in the first round.
This experiment demonstrates that second-round enhancement can further slightly enhance the images or correct over-enhancement.

\vspace{-3.5mm}
\section{Conclusion, Discussion \& Future Work}
\vspace{-1.5mm}
In this paper, we propose a novel quality-guided image enhancement framework that enables the model to associate image features with their perceptual qualities and adjust images continuously according to quality scores.
Initially, to address the lack of a corresponding image retouching quality assessment database, a Skin Tone Image Quality Assessment Database (STIQAD) is established.
We achieve continuous and controllable quality-guided image enhancement on the STIQAD by applying the proposed framework on the 3D LUT method and experimental results also demonstrate that our method achieves better results compared to Image-Adaptive-3DLUT and Seplut.
Moreover, an experiment on 10 natural raw images corroborates the effectiveness of our model in situations with fewer subjects and fewer instances, and also demonstrates that our method is not limited to portrait enhancement.
Extensive additional experiments also demonstrate the generalization and adaptation ability of our method.

The proposed framework can effectively adjust an image with given quality scores, however, some items should be noted and discussed as follows.
First of all, the selection of the images in our STIQAD is careful.
It should be noted that the adjusted images with higher qualities may be in some specific directions relative to the raw image, while the adjusted images with lower qualities may be diverse.
Choosing the adjusted images with higher qualities to cover a continuous range and avoid over-adjusted images can benefit the model to learn continuous positive trends.
Moreover, we also consider the effects of inputting quality scores outside the normalized range during the test process.
We notice that the models at different epochs respond differently to the scores outside the nor range, even though their performance within this range is quite similar. 
Specifically, we apply a score range of [-3, 3] to models that have been trained with scores normalized from -1 to 1. We observe that, in some models, images with scores below -1 become significantly worse, while those with scores above 1 remain similar to 1. 
However, some other models may exhibit an inverse trend where the improvements are amplified, and the degradation remains relatively unchanged.
Finally, it should be noted that the ``quality'' in this work is more related to aesthetic quality rather than signal fidelity.
Each adjusted image in our database is assigned with a quality score.
This score essentially serves as a label that reflects the perceived mean quality of the generated image according to the average aesthetic of most people.
As the quality score is inherently subjective and can vary significantly from person to person, our quality-guided enhancement results may not universally align with every person's perception.
However, as discussed in Section \ref{section:sec5.5}, our pretrained quality guided image enhancement model can be easily trained to adapt to specific opinions.

Our future work will extend the quality-guided framework to cover a broader range of applications.
We will explore the effective objective aesthetic image quality assessment algorithms for adjusted images and apply the corresponding evaluation methods to our quality-guided image enhancement framework.

\vspace{-1.5mm}
\bibliographystyle{IEEEtran}
\vspace{-2.5mm}
\bibliography{arxiv}

\begin{IEEEbiography}[{\includegraphics[width=1in,height=1.25in,clip,keepaspectratio]{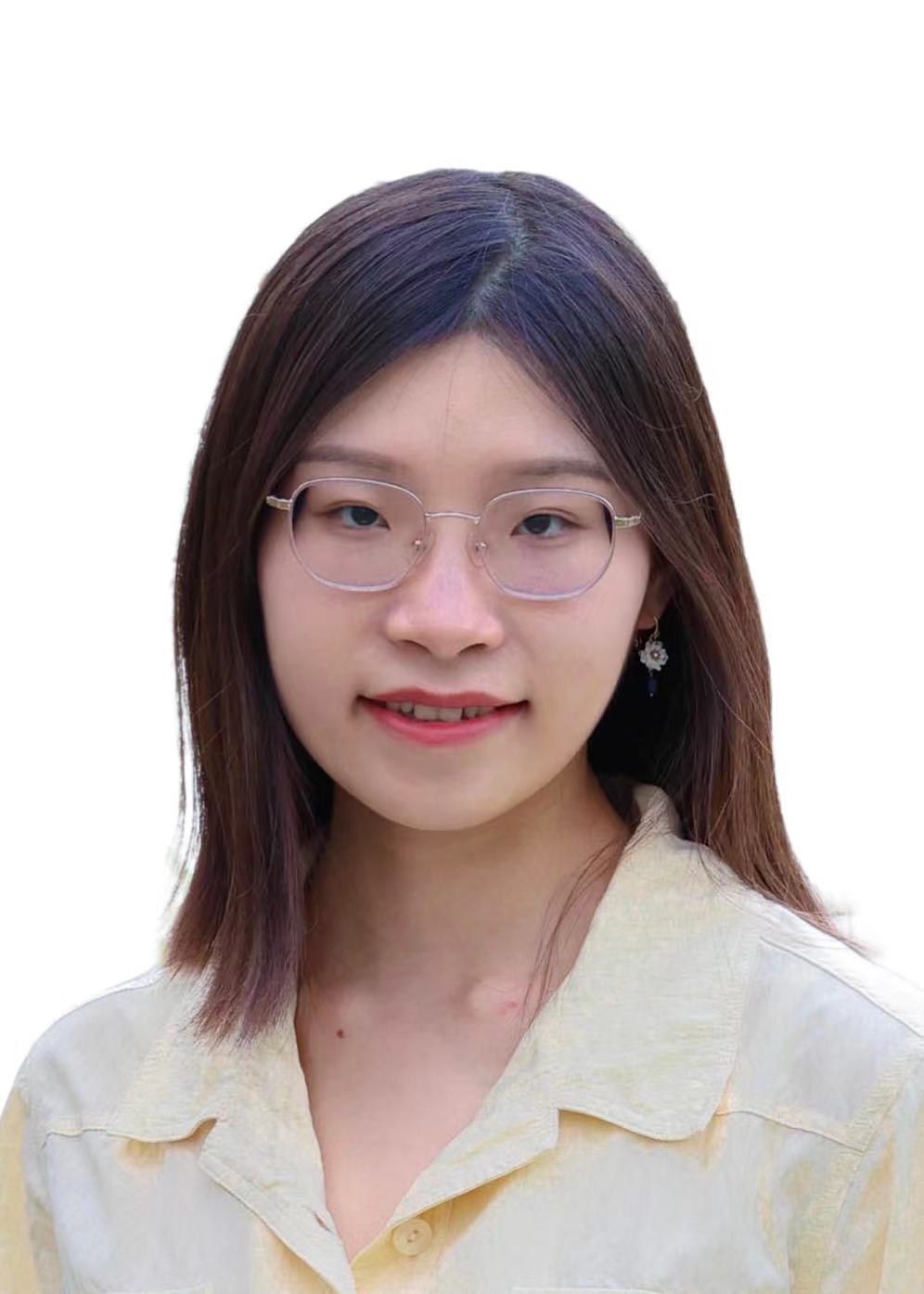}}]{Shiqi Gao}
received the B.E. degree and now pursuing the Ph.D. degree at the Department of Electronic Engineering, Shanghai Jiao Tong University, Shanghai, China. Her research interests include image quality assessment and image enhancement.
\end{IEEEbiography}

\vspace{-12mm}
\begin{IEEEbiography}
[{\includegraphics[width=1in,height=1.25in,clip,keepaspectratio]{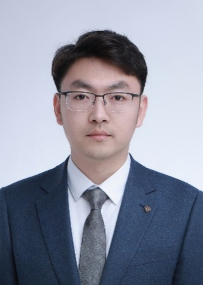}}]{Huiyu Duan}
received the B.E. degree from the University of Electronic Science and Technology of China, Chengdu, China, in 2017. He is currently pursuing the Ph.D. degree with the Department of Electronic Engineering, Shanghai Jiao Tong University, Shanghai, China. From Sept. 2019 to Sept. 2020, he was a visiting Ph.D. student at the Schepens Eye Research Institute, Harvard Medical School, Boston, USA. He received the Best Paper Award of IEEE International Symposium on Broadband Multimedia Systems and Broadcasting (BMSB) in 2022. His research interests include perceptual quality assessment, quality of experience, visual attention modeling, extended reality (XR), and multimedia signal processing.
\end{IEEEbiography}

\vspace{-12mm}
\begin{IEEEbiography}
[{\includegraphics[width=1in,height=1.25in,clip,keepaspectratio]{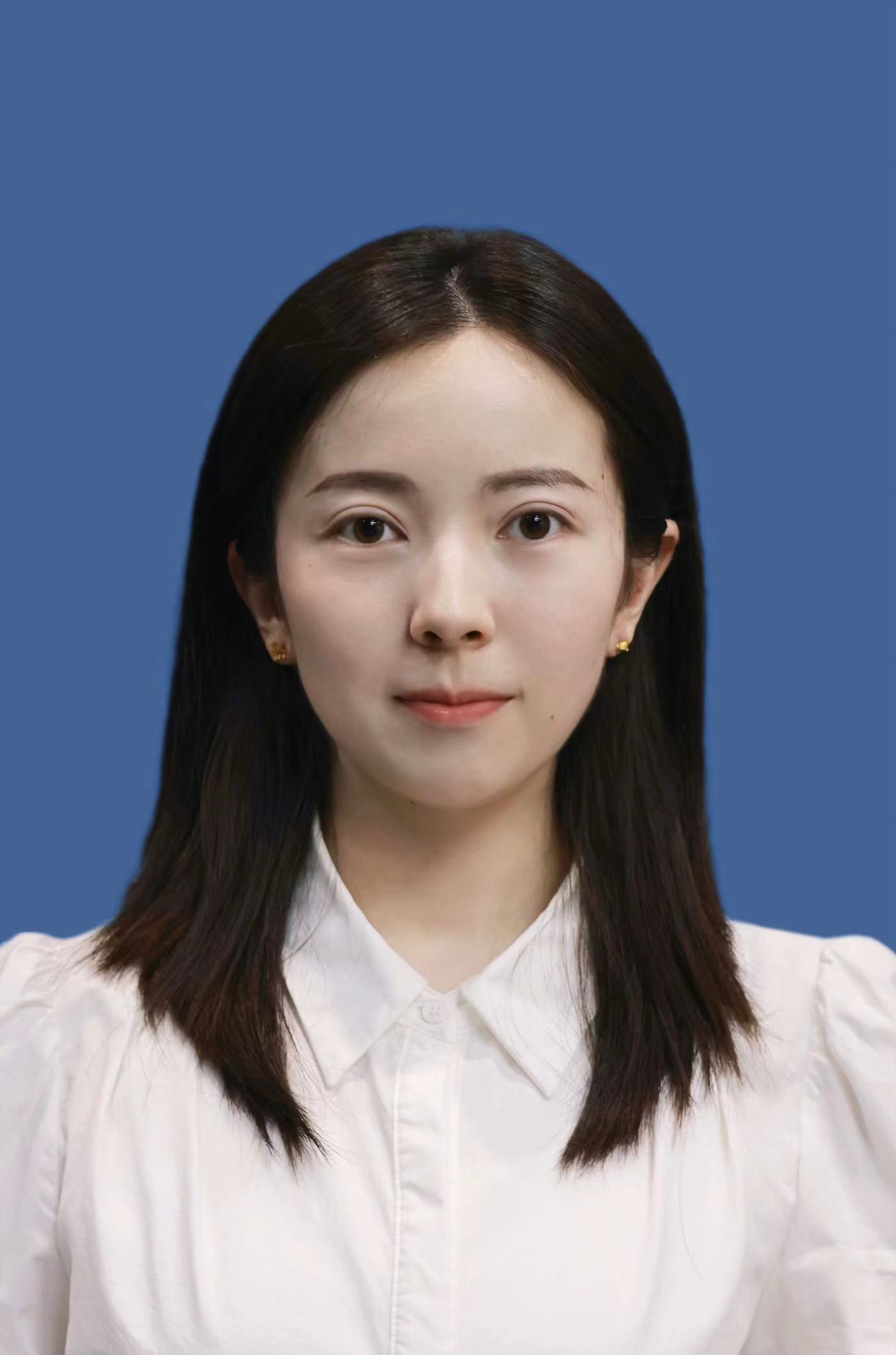}}]{Xinyue Li}
eceived the B.E. and M.E. degree from the Shanghai Ocean University of China, Shanghai, China, in 2023. She is currently pursuing the Ph.D. degree with the Department of Electronic Engineering, Shanghai Jiao Tong University, Shanghai, China. From Sept. Her research interests include multimedia signal processing and AI for EDA.
\end{IEEEbiography}

\vspace{-12mm}
\begin{IEEEbiography}
[{\includegraphics[width=1in,height=1.25in,clip,keepaspectratio]{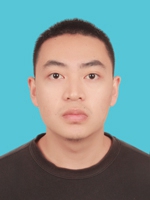}}]{Kang Fu}
received his B.E. degree from Shanghai University, Shanghai, China, in 2020 and he is currently pursuing his Ph.D. degree at the School of Electronic Information and Electrical Engineering at Shanghai Jiao Tong University. His research interests include image quality assessment, video quality assessment, and 3D visual quality assessment.
\end{IEEEbiography}

\vspace{-12mm}
\begin{IEEEbiography}
[{\includegraphics[width=1in,height=1.25in,clip,keepaspectratio]{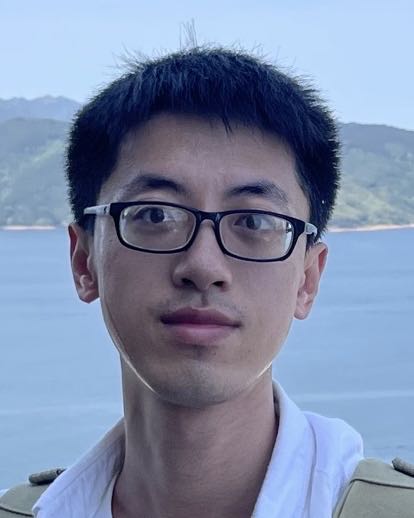}}]{Yicong Peng}
received the B.E. degree and now working toward the Ph.D. degree in Information and Communication Engineering from Shanghai Jiao Tong University, Shanghai, China. His current research interests include computer vision, 3D rendereing, digital human and generative artifitial intelligence.
\end{IEEEbiography}

\vspace{-12mm}
\begin{IEEEbiography}
[{\includegraphics[width=1in,height=1.25in,clip,keepaspectratio]{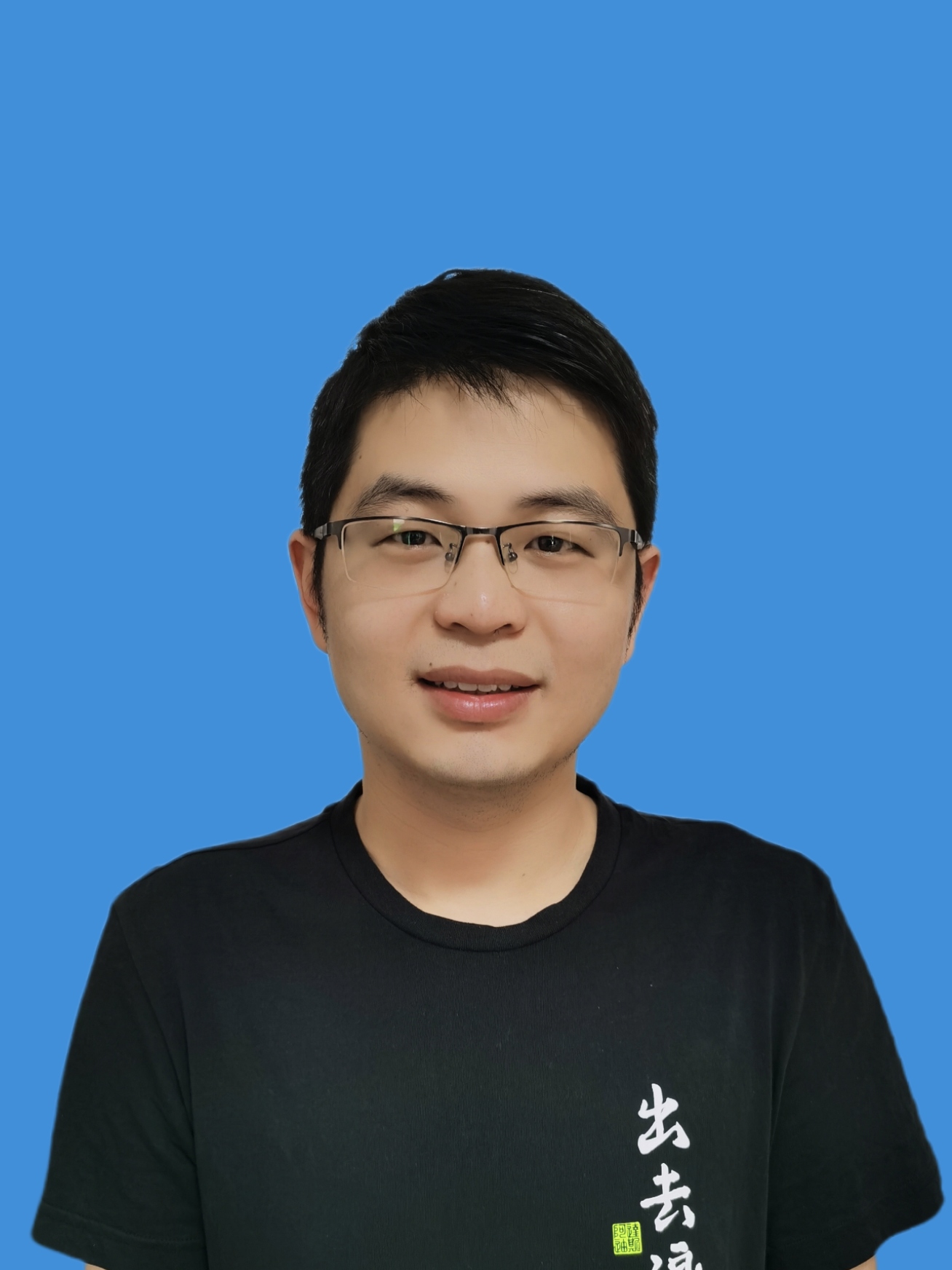}}]{Qihang Xu}
,an image algorithm engineer at Transsion. His current research interests are portrait perception, image reatouching and image editing.
\end{IEEEbiography}

\vspace{-12mm}
\begin{IEEEbiography}
[{\includegraphics[width=1in,height=1.25in,clip,keepaspectratio]{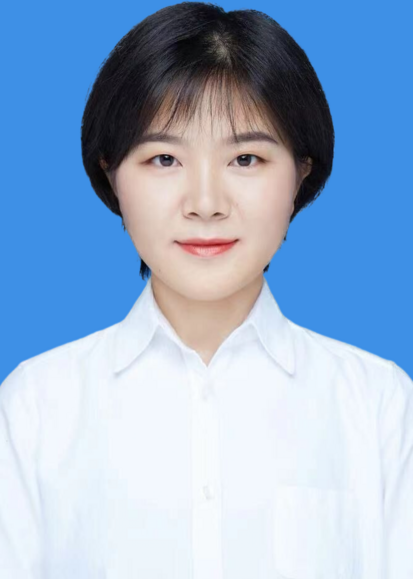}}]{Yuanyuan Chang}
,an image algorithm engineer at Transsion. Her current research interests are color science, image reatouching and image editing.
\end{IEEEbiography}

\vspace{-12mm}
\begin{IEEEbiography}
[{\includegraphics[width=1in,height=1.25in,clip,keepaspectratio]{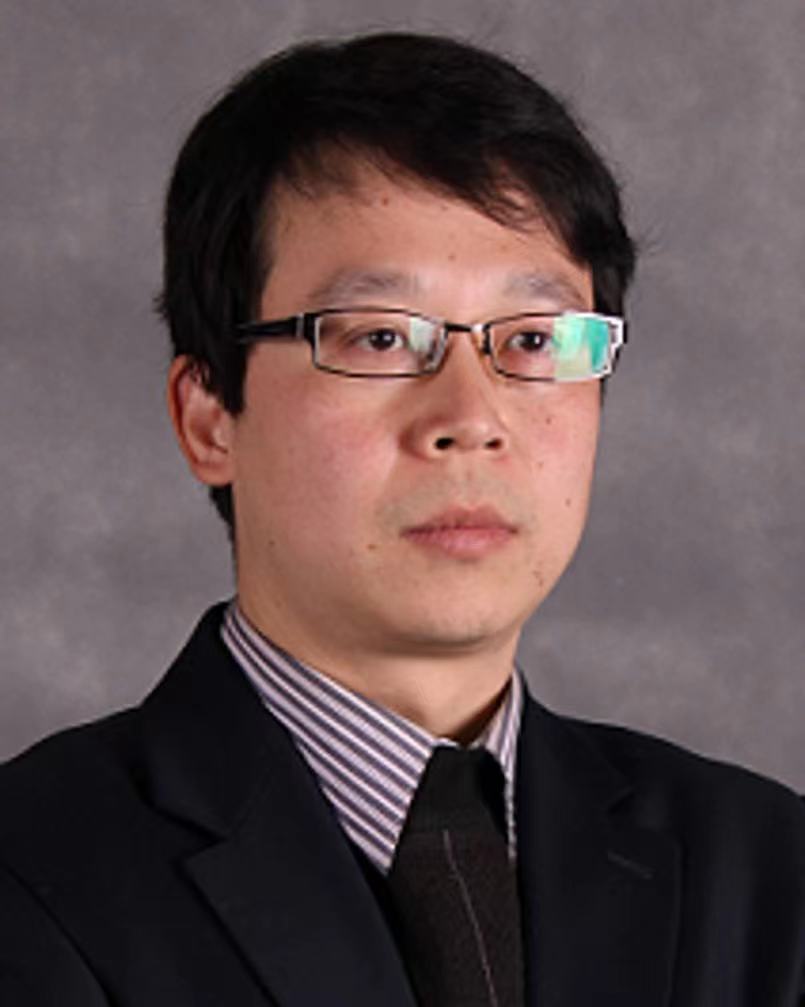}}]{Jia Wang}
received the Ph.D. degree in electronic engineering from Shanghai Jiao Tong University, China in 2002. He is currently a Professor with the Department of Electronics Engineering, Shanghai Jiao Tong University. His research interests include multiuser information theory and mathematics in artificial intelligence. He is a member of Shanghai Key Laboratory of Digital Media Processing and Transmission.
\end{IEEEbiography}

\vspace{-90mm}
\begin{IEEEbiography}
[{\includegraphics[width=1in,height=1.25in,clip,keepaspectratio]{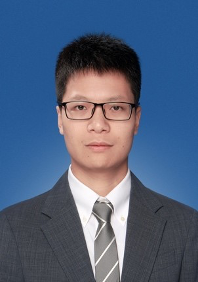}}]{Xiongkuo Min}
received the B.E. degree from Wuhan University, Wuhan, China, in 2013, and the Ph.D. degree from Shanghai Jiao Tong University, Shanghai, China, in 2018, where he is currently a tenure-track Associate Professor with the Institute of Image Communication and Network Engineering. From Jan. 2016 to Jan. 2017, he was a visiting student at University of Waterloo. From Jun. 2018 to Sept. 2021, he was a Postdoc at Shanghai Jiao Tong University. From Jan. 2019 to Jan. 2021, he was a visiting Postdoc at The University of Texas at Austin. He received the Best Paper Runner-up Award of IEEE Transactions on Multimedia in 2021, the Best Student Paper Award of IEEE International Conference on Multimedia and Expo (ICME) in 2016, and the excellent Ph.D. thesis award from the Chinese Institute of Electronics (CIE) in 2020. His research interests include image/video/audio quality assessment, quality of experience, visual attention modeling, extended reality, and multimodal signal processing.
\end{IEEEbiography}

\vspace{-95mm}
\begin{IEEEbiography}
[{\includegraphics[width=1in,height=1.25in,clip,keepaspectratio]{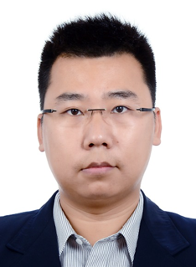}}]{Guangtao Zhai (SM’19)}
received the B.E. and M.E. degrees from Shandong University, Shandong, China, in 2001 and 2004, respectively, and the Ph.D. degree from Shanghai Jiao Tong University, Shanghai, China, in 2009, where he is currently a Research Professor with the Institute of Image Communication and Information Processing. From 2008 to 2009, he was a Visiting Student with the Department of Electrical and Computer Engineering, McMaster University, Hamilton, ON, Canada, where he was a Post-Doctoral Fellow from 2010 to 2012. From 2012 to 2013, he was a Humboldt Research Fellow with the Institute of Multimedia Communication and Signal Processing, Friedrich Alexander University of Erlangen-Nuremberg, Germany. He received the Award of National Excellent Ph.D. Thesis from the Ministry of Education of China in 2012. His research interests include multimedia signal processing and perceptual signal processing.
\end{IEEEbiography}

\end{document}